\documentclass[10pt,twocolumn,letterpaper]{article}

\usepackage[pagenumbers,algorithms]{wacv}

\usepackage{wacv}
\usepackage{times}
\usepackage{epsfig}
\usepackage{graphicx}
\usepackage{amsmath}
\usepackage{amssymb}
\usepackage{booktabs}
\usepackage{dsfont}
\usepackage{algorithm2e}
\usepackage{bbm}
\usepackage{listings}
\usepackage{xcolor}
\usepackage{caption}
\usepackage{subcaption}
\usepackage[all]{nowidow}
\usepackage{amsthm}
\usepackage{enumitem}
\usepackage[accsupp]{axessibility}
\usepackage{multirow}

\usepackage[symbol]{footmisc}

\usepackage{amsmath,amsfonts,bm}

\def\eqref#1{equation~\ref{#1}}

\def\1{\bm{1}}

\def\rvh{{\mathbf{h}}}

\def\rvk{{\mathbf{k}}}

\def\rvs{{\mathbf{s}}}

\def\rvw{{\mathbf{w}}}

\def\rvy{{\mathbf{y}}}
\def\rvz{{\mathbf{z}}}

\def\mP{{\bm{P}}}
\def\mQ{{\bm{Q}}}

\def\mX{{\bm{X}}}

\DeclareMathAlphabet{\mathsfit}{\encodingdefault}{\sfdefault}{m}{sl}
\SetMathAlphabet{\mathsfit}{bold}{\encodingdefault}{\sfdefault}{bx}{n}

\def\sR{{\mathbb{R}}}

\def\wacvPaperID{6} 
\def\confName{WACV}
\def\confYear{2024}

\usepackage[pagebackref,breaklinks,colorlinks]{hyperref}

\usepackage[capitalize]{cleveref}
\crefname{section}{Sec.}{Secs.}
\Crefname{section}{Section}{Sections}
\Crefname{table}{Table}{Tables}
\crefname{table}{Tab.}{Tabs.}

\newtheorem*{prop*}{Proposition}

\SetKwComment{Comment}{/* }{ */}
\SetKwInput{KwData}{Input}

\graphicspath{{images/}}

\definecolor{codegreen}{rgb}{0,0.6,0}
\definecolor{codegray}{rgb}{0.5,0.5,0.5}
\definecolor{codepurple}{rgb}{0.58,0,0.82}
\definecolor{backcolour}{rgb}{0.95,0.95,0.92}

\lstdefinestyle{mystyle}{
    backgroundcolor=\color{backcolour},   
    commentstyle=\color{codegreen},
    keywordstyle=\color{magenta},
    numberstyle=\tiny\color{codegray},
    stringstyle=\color{codepurple},
    basicstyle=\ttfamily\footnotesize,
    breakatwhitespace=false,         
    breaklines=true,                 
    captionpos=b,                    
    keepspaces=true,                 
    numbers=left,                    
    numbersep=5pt,                  
    showspaces=false,                
    showstringspaces=false,
    showtabs=false,                  
    tabsize=2,
    showlines=true
}

\begin{document}

\title{COMEDIAN: Self-Supervised Learning and Knowledge Distillation for Action Spotting using Transformers}

\author{
Julien Denize\,\footnote[1]{} \footnote[2]{} \and Mykola Liashuha\,\footnote[1]{} \and Jaonary Rabarisoa\,\footnote[1]{} \and Astrid Orcesi\,\footnote[1]{} \and Romain Hérault\,\footnote[2]{} \\
\footnote[1]{} \,  Université Paris-Saclay, CEA, LIST, F-91120, Palaiseau, France \\
{\tt\small firstname.lastname@cea.fr} \\
\footnote[2]{} \, Normandie Univ, INSA Rouen, LITIS, 76801, Saint Etienne du Rouvray, France \\
{\tt\small firstname.lastname@insa-rouen.fr}
}
\maketitle

\begin{abstract}
    We present COMEDIAN, a novel pipeline to initialize spatiotemporal transformers for action spotting, which involves self-supervised learning and knowledge distillation. Action spotting is a timestamp-level temporal action detection task. Our pipeline consists of three steps, with two initialization stages. First, we perform self-supervised initialization of a spatial transformer using short videos as input. Additionally, we initialize a temporal transformer that enhances the spatial transformer's outputs with global context through knowledge distillation from a pre-computed feature bank aligned with each short video segment. In the final step, we fine-tune the transformers to the action spotting task. The experiments, conducted on the SoccerNet-v2 dataset, demonstrate state-of-the-art performance and validate the effectiveness of COMEDIAN's pretraining paradigm. Our results highlight several advantages of our pretraining pipeline, including improved performance and faster convergence compared to non-pretrained models. Source code is available here: \url{https://github.com/juliendenize/eztorch}.
\end{abstract}

\begin{figure*}[t]
\centering
\includegraphics[trim=0cm 0cm 22cm 0cm, angle=-90,origin=c,width=0.9\textwidth]{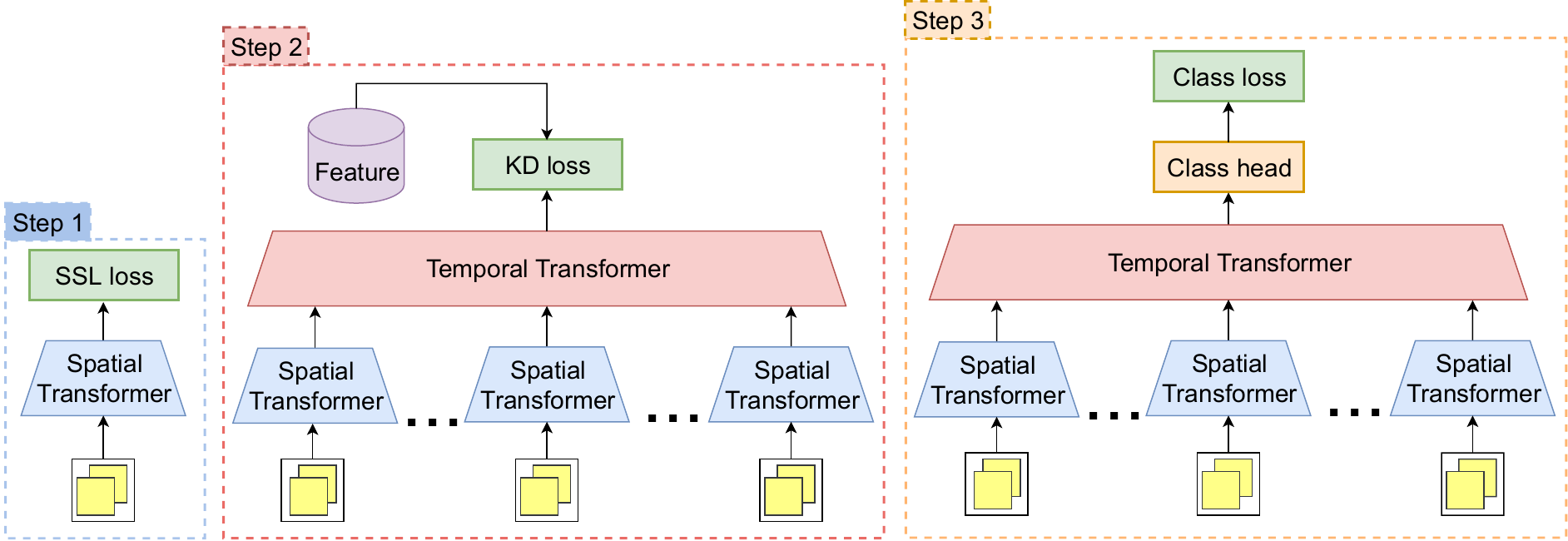}
\caption{Overview of COMEDIAN training pipeline. Step 1: Pretraining of the spatial transformer. Step 2: Pretraining of the spatial and temporal transformers. Step 3: fine-tuning to the action spotting task.}
\label{fig:overview}
\end{figure*}

\section{Introduction}

The field of computer vision has witnessed remarkable progress in recent years, and in particular in video analysis since the Deep Learning era. To make the best use of Deep Learning models, vast amounts of data and high computational power are needed. Temporal Action Detection (TAD) has been gaining attention as it has different applications in our lives ranging from home automation to sports analysis. TAD involves the identification of when specific actions occur within a video, enabling comprehensive understanding and meaningful insights into the dynamics of a given scenario. Action Spotting \cite{giancola2018} is a specific TAD task whose goal is to predict actions at a precise timestamp and therefore requires a temporally precise prediction. Modeling actions in videos faces several issues such as the sparsity of actions and the intricate relationships between them.

While a lot of video data is readily available, annotating actions presents significant challenges. It suffers from the inherent subjectivity of annotators to interpret when an action starts or ends. Moreover, the process of manual annotation requires considerable time and resources, limiting the scalability of annotating large datasets. 

Recent advancements in vision Transformers \cite{dosovitskiy2021, liu2021} for video analysis \cite{arnab2021, liu2022b, selva2023} showed them surpassing the traditionally used Convolutional Neural Networks (CNNs) but required dedicated architecture design to reduce computational cost \cite{neimark2021, arnab2021, bertasisus2021, xu2021, wu2022, yang2022, meinhardt2022}. By capturing long-range dependencies and leveraging global context, Transformers analyses better complex sequences. However, their effectiveness crucially depends on proper initialization and access to ample labeled training data \cite{arnab2021, liu2022b}.

The use of pretraining methods has shown tremendous progress in enhancing the capabilities of Transformers \cite{arnab2021, caron2021, liu2022b}. Pretraining can be achieved through two approaches: Supervised Learning (SL) using labeled data that requires extensive annotation efforts, and Self-Supervised Learning (SSL) that leverages unlabeled data. SSL for video representation learning \cite{Feichtenhofer2021, tong2022, wei2022, denize2023b} has shown promising results, demonstrating significant improvements in generalization and narrowing the gap with supervised learning approaches. Knowledge Distillation (KD) \cite{hinton2015} also serves as a powerful tool to initialize a network by transferring knowledge from another network or a collection of models. Depending on how the networks are obtained and the losses used to distill, KD can be considered as SL \cite{touvron2021} or SSL \cite{gao2022}.

In this work, we propose the COMEDIAN approach, which combines self-supervised learning and knowledge distillation to initialize a spatio-temporal transformer for action spotting. Our method involves two transformers: a spatial transformer, which learns short context information from frames extracted from small videos, and a temporal transformer, which enriches the local context with global information. The initialization process consists of two stages: the first stage focuses on the spatial transformer via SSL, while the second stage initializes both spatial and temporal transformers via KD. KD is employed from a pre-computed bank of representations aligned with each output temporal token. Importantly, COMEDIAN leverages unlabeled video data for initialization, effectively addressing the aforementioned challenges associated with transformers.

To evaluate our approach, we conducted experiments on the action spotting task on the SoccerNet-v2 \cite{deliege2021} dataset, which contains soccer matches with 17 distinct actions varying in semantics and occurrence. Our contributions can be summarized as follows:
\begin{itemize}[noitemsep,topsep=0pt,leftmargin=*]
    \item We propose COMEDIAN a combined self-supervised learning and knowledge distillation pipeline illustrated in \cref{fig:overview} to initialize transformers for action spotting.
    \item We demonstrate that COMEDIAN achieves state-of-the-art performance for action spotting on the SoccerNet-v2 dataset, showcasing the effectiveness of our self-supervised and knowledge distillation pipeline. 
    \item We provide a comprehensive analysis of the benefits of pretraining with knowledge distillation, including improved performance and faster convergence compared to non-pretrained models.
\end{itemize}

\section{Related Work}

\textbf{Video Transformers.} Vision Transformers \cite{dosovitskiy2021} (ViT) capture long-term dependencies better than recurrent models or convolutional networks. It relies on a tokenizer, that embeds patches of the input, and self-attention \cite{vaswani2017}. Standard self-attention is computationally heavy and several other attention mechanisms have been proposed such as Swin \cite{liu2021} or DeIT \cite{touvron2021}. Transformers can be applied to videos by adapting the tokenizer \cite{arnab2021, bertasisus2021, liu2022b}. However as videos increase the number of tokens, video transformers are computationally heavy, and various strategies have been proposed to reduce their cost. VTN \cite{neimark2021} adds a temporal encoder on top of a ViT while ViViT \cite{arnab2021} and TimeSformer \cite{bertasisus2021} propose several factorizations of space-time attention. ViViT as VTN found a spatio-temporal hierarchical model offers the best trade-off between performance and cost which led to several methods \cite{bulat2021, hwang2021}. Previously mentioned transformers focused on short videos, e.g. $\le 5$ seconds and some architectures have been developed to capture long-range dependencies on longer videos via a sliding window that keeps relevant information from the past with a memory \cite{xu2021,wu2022} or a recurrence \cite{yang2022, meinhardt2022} mechanism. 

In our work, we consider a spatio-temporal hierarchical model without changing specifically the architecture to capture long-term dependencies as we also want to capture bidirectional short-term dependencies.

\textbf{Pretraining.} Pretraining has been crucial to unleashing image \cite{dosovitskiy2021} and video \cite{arnab2021} transformers either via Supervised Learning (SL) on a large dataset \cite{dosovitskiy2021} or Self-Supervised Learning (SSL) \cite{chen2021b, caron2021}. Contrastive learning \cite{VanDenOord2018} is a kind of state-of-the-art SSL on images \cite{He2020, Chen2020b, Caron2020, Grill2020, Dwibedi2021, denize2023} that has been adapted to video \cite{Lorre2020, Qian2021, Feichtenhofer2021, Recasens2021, Dave2022, denize2023b}. It pulls representations of positive views based on the input while pushing a large number of other representations. Recently, SCE \cite{denize2023b} has shown that leveraging estimated inter-instance relations with contrastive learning improves performance. Masked Modeling approaches emerged with transformers and showed better performance than contrastive learning to learn local features \cite{he2022, bao2022, zhou2021b, oquab2023}. It masks a part of the input and reconstructs the signal either at pixel-level like MAE \cite{he2022, tong2022, feichtenhofer2022}, at features-level \cite{wei2022,gao2022}, or by predicting visual tokens \cite{sun2019b, wang2022}. However, these methods assume lots of redundancies are present in the video which is true for short videos with few view variations but does not hold for complex videos such as soccer matches.  Finally, Knowledge distillation (KD) \cite{hinton2015} is another pretraining approach that distills information from a teacher or a collection of teacher models to a student and has been successfully applied for Supervised Learning \cite{touvron2021} as well as Self-Supervised Learning \cite{park2019, fang2021, koohpayegani2020, gao2022}. Multiple approaches emerged to learn spatial and temporal features separately or decoupled to learn a global representation \cite{huang2021a, qian2022}. Notably, \cite{zhou2022a} proposes a multi-step pretraining method that decouples spatial and temporal information through two different networks recoupled via a self-distilled network. \cite{zhang2022} performs spatial and temporal contrastive learning at multiple hierarchies in the model to separate spatial and temporal features. 

In our work, we  pretrain our hierarchical model with a contrastive SSL initialization of the spatial transformer. Then, our global model is pretrained with a KD loss from an extracted bank of features aligned with all the output tokens. This loss leverages temporal masking and soft contrastive learning to maintain local-temporal information enriched in a global context. Therefore, our goal is to learn multiple local-temporal representations, not one global. 

\textbf{Action Spotting.} Action Spotting is a timestamp-level Temporal Action Detection (TAD) first introduced for the dataset SoccerNet \cite{giancola2018}. This dataset has been extended to more videos and more actions in SoccerNet-v2 \cite{deliege2021} and the tight-Average mean Average Precision (t-AmAP) has been introduced to evaluate precise detection within thresholds of 1 to 5 seconds. Several approaches have been proposed to tackle this task that can be divided into two categories. First, most approaches build a temporal architecture on top of a feature extractor \cite{cioppa2020, tomei2020, minoura2021, zhou2021, mahaseni2021, giancola2021, shi2022, chen2022, cao2022, darwish2022, soares2022}, and second, few others train an end-to-end network \cite{zhu2022, hong2022}. The first kind of approach reduces the computational cost of experiments however makes them rely on a feature extractor for generalization. Notably, Baidu \cite{zhou2021} proposed the extraction features of five 3D CNN models pretrained on Kinetics \cite{kakogeorgiou2022} and finetuned on SoccerNet-v2. The features are then plugged into a temporal action detector. Spivak \cite{soares2022} used these Baidu features coupled with features from a pretrained ResNet-152 to train an anchor-based approach that first classifies actions falling in a few-second temporal radius and shifts the predictions using a temporal regressor. For end-to-end approaches, E2E-spot \cite{hong2022} proposed to train a CNN spatial encoder on top of a simple recurrent model that performed competitively with the previous approach. As for other domains, the attention mechanism has been studied to improve performance \cite{minoura2021,zhou2021,zhu2022,shi2022,soares2022}.

In our work, we propose an end-to-end transformer-based action spotting approach that assigns actions that fall in frames in a small temporal radius.

\section{Method}

\subsection{Overview}
Our approach seeks to learn locally precise temporal features enriched with a larger context for action spotting. Therefore our model is composed of three different parts:
\begin{itemize}[noitemsep,topsep=0pt,leftmargin=*]
    \item A spatial transformer embeds the information of a small temporal window and outputs one token embedding.
    \item A temporal transformer that takes as input the token outputs of the spatial transformer on consecutive windows and outputs the same number of tokens. The temporal transformer enriches the representation of local windows with the knowledge of a larger context.
    \item A linear head is applied on each temporal output token to perform the classification of the action classes.
\end{itemize}

To train the model, we perform three different training steps described in the latter subsections: (1) pretraining of the spatial transformer on small windows in \cref{subsec:spatial}, (2) pretraining of the spatial and temporal transformers on large windows in \cref{subsec:spatio-temporal}, (3) fine-tuning of the model on the action spotting task in \cref{subsec:fine-tuning}.

\subsection{Spatial pretraining}\label{subsec:spatial}

\begin{figure}[t]
\centering
\includegraphics[width=0.25\textwidth]{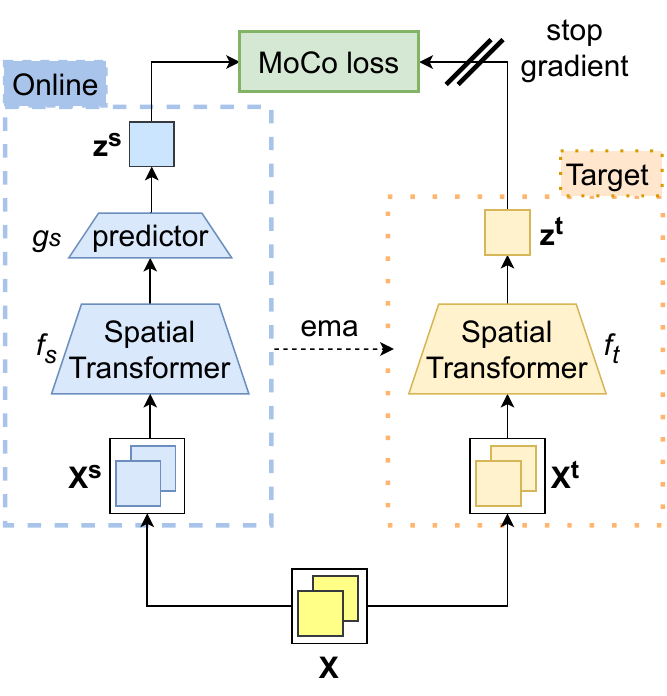}
\caption{Pretraining of the spatial transformer.}
\label{fig:spatial}
\end{figure}

To train the spatial transformer, we follow the Self-Supervised Contrastive method MoCo \cite{He2020} illustrated in \cref{fig:spatial}. We use a Siamese architecture containing an online and a target branch. For the online branch, the model contains a transformer $f_s$ and a predictor $g_s$. The target branch contains a copy of the transformer updated by the exponential moving average, or ema, of the online transformer.

Each video $\mX \in \sR^{T_s \times H \times W \times C}$ of $T_s$ frames, width $W$, height $H$, and $C$ channels from the dataset is augmented by two different distributions of data augmentations $A^1$ and $A^2$ to form positive views $\mX^1 = a^1(\mX)$, $\mX^2 = a^2(\mX)$ with $a^1 \sim A^1$ and $a^2 \sim A^2$. We pass both views in both transformers to compute the representations $\rvz^{1s} = g_s(f_s(\mX^1))$, $\rvz^{2s} = g_s(f_s(\mX^2))$, $\rvz^{1t} = f_t(\mX^1)$ and $\rvz^{2t} = f_t(\mX^2)$. A momentum memory buffer $\mQ$ of size $M >> N$ is maintained on the target representations to provide negatives.

We apply the MoCo loss on each representation as follows:
\begin{align}
& \mathcal{L}_{MoCo} =  -\frac{1}{2} \left( l_M(\rvz^{1s}, \rvz^{2t}) + l_M(\rvz^{2s}, \rvz^{1t}) \right), \\
& l_M(\rvz, \rvk) = log \left(\frac{exp(\rvz \cdot \rvk / \tau)}{exp(\rvz \cdot \rvk / \tau) + \sum_{j=1}^nexp(\rvz \cdot \mQ_j / \tau)} \right).
\end{align}

\subsection{Spatio-temporal pretraining}\label{subsec:spatio-temporal}

\begin{figure}[t]
\centering
\includegraphics[trim=0 2.9cm 0 0,clip, width=0.45\textwidth]{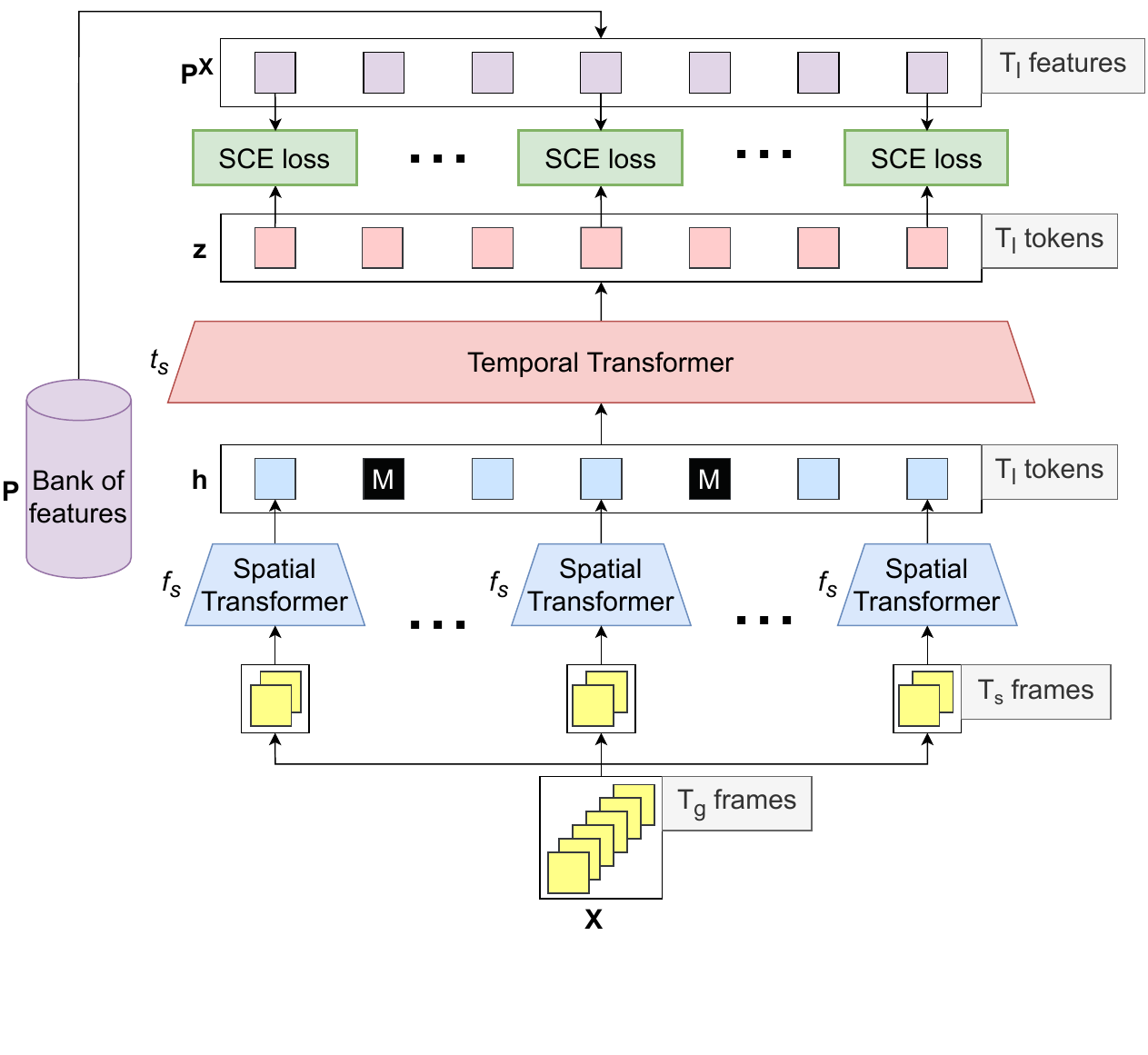}
\caption{Pretraining of the spatial and temporal transformers via knowledge distillation of a bank of features with the SCE loss. Some spatial output tokens are masked.}
\label{fig:spatial-temporal}
\end{figure}

To pretrain the spatial transformer $f_s$ and the temporal transformer $t_s$, we adapt the Soft Contrastive Self-Supervised loss SCE to perform knowledge distillation as illustrated in \cref{fig:spatial-temporal}. To do so, we consider the following inputs:
\begin{itemize}[noitemsep,topsep=0pt,leftmargin=*]
    \item A large video $\mX \in \sR^{T_g, C, H, W}$ of $T_g$ frames that is divisible by $T_s$ with $T_l = T_g / T_s$ sampled from a dataset. 
    \item A bank of spatio-temporal features $\mP \in \sR^{M_P, D_t}$ of size $M_P$ and dimension $D_t$ that can be aligned temporally with small window of $T_s$ frames within any sampled large video. More specifically, each small window is associated with the temporally closest feature of its middle frame. Therefore for each sample $\mX$ a set of features $\mP^\mX \in \sR^{T_l, D_t}$ is selected from $\mP$. As a preprocessing stage, this bank of features is extracted from a pretrained model.
\end{itemize}

The video is augmented with $a^{3} \sim T^{3}$ such as $\mX^{3} = a^{3}(\mX)$. Each global window is split into $T_l = T_g / T_s$ smaller windows and the input is reshaped to $(T_l, T_s, C, H, W)$. It passes through the spatial transformer to output $\rvh = f_s(\mX^{3})$ with $\rvh \in \sR^{T_l \times D}$ and $D$ the output dimension of the spatial transformer. A masking ratio $\alpha_1$ is applied to replace $\alpha_1 * T_l$ tokens with a learned mask token.  The temporal transformer adds a positional embedding to each token and computes $\rvz = t_s(\rvh)$ with $\rvz \in \sR^{T_l, D_t}$ and $D_t$ the dimension of the temporal output tokens.

The SCE loss is applied on each token of $\rvz$ with the associated set of features $\mP^\mX$ as follows. First, a target relation distribution $\rvs^2$ is computed on each of the features in $\mP^\mX$ with the complete bank. A one-hot label is mixed with this distribution with a coefficient $\lambda$ to form the target distribution $\rvw^2$. Then, each token in $\rvh$ predicts this target distribution by computing its similarity distribution $\rvs^1$ with the complete bank of features. Finally, the loss is applied:

\begin{align}
    & s^{1}_{ik} = \frac{\exp(\rvz^1_i \cdot \mP_{k} / \tau)}{\sum_{j=1}^{M_P}{\cdot \exp(\rvz^1_i \cdot \mP_j / \tau)}}, \\
    & s^{2}_{ik} = \frac{\mathbbm{1}_{\mP^\mX_i \neq \mP_k} \cdot \exp(\mP^\mX_{i} \cdot \mP_k / \tau_m)}{\sum_{j=1}^{M_P}{\mathbbm{1}}_{\mP^\mX_i \neq \mP_j} \cdot \exp(\mP^\mX_i \cdot \mP_j / \tau_m)}, \\
    & w^{2}_{ik} = \lambda \mathbbm{1}_{\mP^\mX_i = \mP_k} + (1 - \lambda) s^{2}_{ik}, \\
    & \mathcal{L}_{SCE} = - \frac{1}{T_l}\sum_{i = 1}^{T_l}\rvw^2_i\log(\rvs^1_i).
\end{align}

The KD enforces that each temporal token contains the information of its corresponding smaller window while allowing contextual information from the larger window thanks to the temporal transformer. The SCE loss enables the spatio-temporal token representation to model the relations among spatio-temporal small windows that the bank of features contains. Depending on how the features are extracted, the KD can be considered as supervised or self-supervised as discussed in \cref{subsec:ablation}.

\subsection{Fine-tuning}\label{subsec:fine-tuning}

\begin{figure}[t]
\centering
\includegraphics[trim=0 2.9cm 0 0,clip, width=0.4\textwidth]{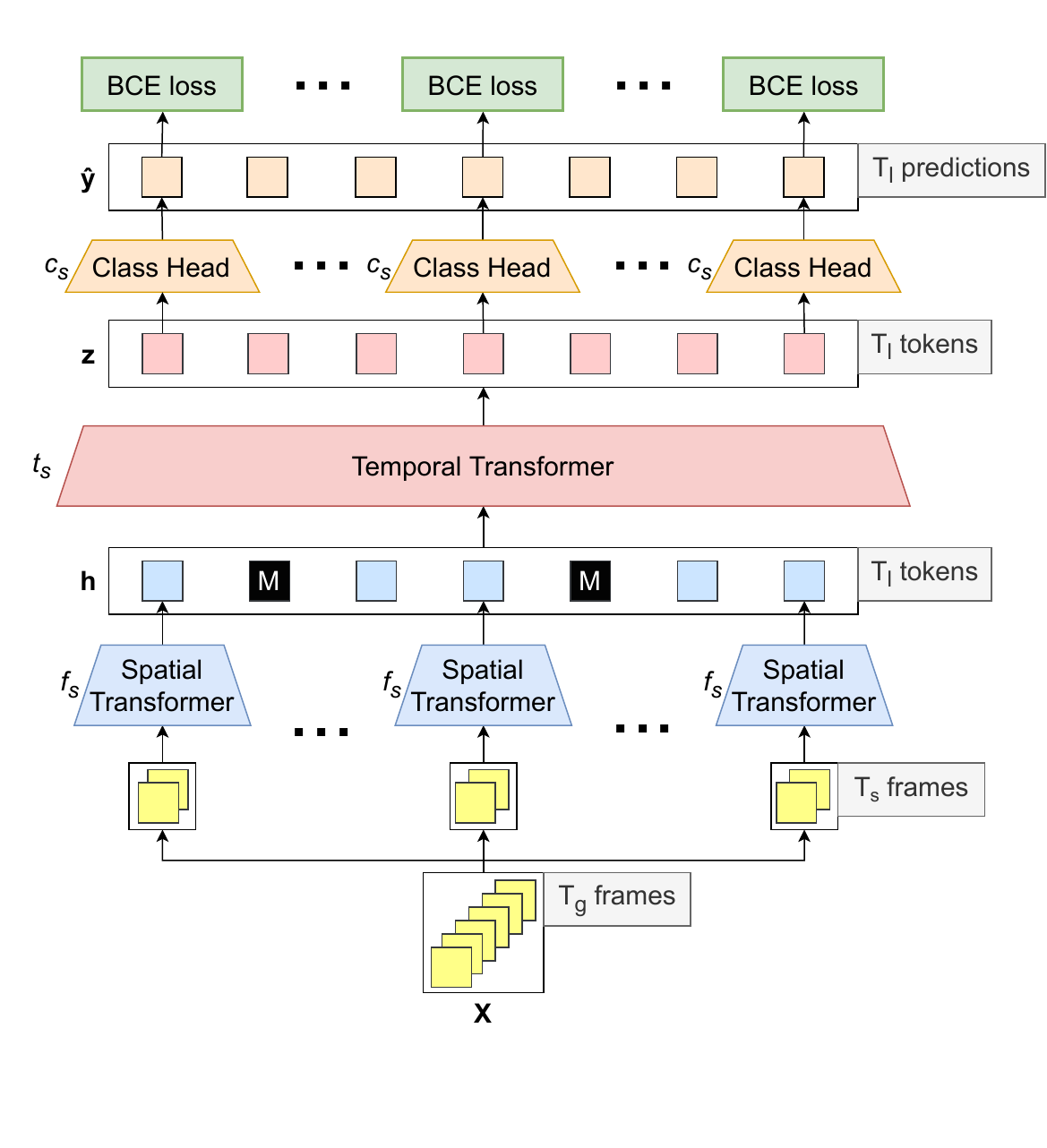}
\caption{Fine-tuning to the action spotting task. Some spatial output tokens are masked.}
\label{fig:fine-tuning}
\end{figure}

The input video $\mX$ considered in this section has the same shape as for spatio-temporal pretraining and passes through the spatial and temporal transformer that outputs $\rvz = t_s(f_s(a^4(\mX)))$ with $a^4 \sim A^4$, a data augmentation.

To train the model to the action spotting task with $C$ classes, a classification head $c_s$ is placed upon the temporal transformer and is applied on each of the temporal output tokens to predict $\hat{\rvy} = c_s(\rvz)$ with $\hat{\rvy} \in \sR^{T_l, C}$ as illustrated in \cref{fig:fine-tuning}. Each token is associated with the average timestamp to the corresponding small window they represent. The masking strategy from the spatiotemporal pretraining is maintained during training by randomly masking $\alpha_2 \times T_l$ spatial output tokens.

For supervision, each ground truth action that falls into the $T_g$ sampled window is associated with the input frame timestamps that fall into a temporal radius displacement $\epsilon$ with the action. For each $T_s$ smaller window, if at least one of its frames is associated with an action, the label associated with its temporal output token is 1 and otherwise 0 to form the vector of label $\rvy \in \sR^{T_l, C}$.

The action classes are considered independently, therefore we apply a Sigmoid activation function to the classifier. During training, the Binary Cross Entropy (BCE) loss is computed for each class at each timestamp as follows:

\begin{equation}
    \mathcal{L}_{BCE} = - \frac{1}{T_l \times C} \sum_{t=1}^{T_l}\sum_{c=1}^{C} \rvy_{tc}log(\hat{\rvy}_{tc})
\end{equation}

During inference, the predictions are assigned to the output timestamps of their temporal token. A sliding window with overlap is performed on the videos. For overlapped predictions, a strategy is applied to only keep one prediction per class per timestamp, such as keeping the maximum or the average of predictions.

\section{Empirical study}

In this section, we will first review the implementation details of each step, then perform an ablation study on the different parts of our pipeline, and finally compare ourselves with the state of the art.

\subsection{Implementation details}

We launched our experiments on three seeds and averaged the results.

\textbf{Dataset.} We performed our study on the SoccerNet-v2 \cite{deliege2021} action spotting dataset. It contains soccer matches divided into two halves of about 45 minutes. It has three annotated splits of 17 classes with 300 matches for training, 100 for validation, and 100 for testing. There is also a challenge split that contains 50 videos for which annotations are not given. The metric used is the tight-Average mean Average Precision, t-AmAP for short, which evaluates predictions that fall on average between 1-5 seconds. We extracted the video frames at resolution $398 \times 224$ at 2 Frames Per Second (FPS). The dataset provides pre-computed Baidu \cite{zhou2021} features at 1 FPS for all splits. We performed a PCA on these features to reduce the dimension to $512$ for distillation.

\textbf{Spatial and Temporal transformer architectures.} For the spatial transformer, two different architectures are used: ViT \cite{dosovitskiy2021} and Swin \cite{liu2021}. The temporal transformer is a stack of 4 attention layers from a ViT architecture. For ViT, the global architecture corresponds to the ViViT model 2 \cite{arnab2021}. We keep this name and refer to the Swin-based architecture as ViSwin. More details can be found in the supplementary material. For optimization, we used the ADAMW optimizer with a weight decay of $0.05$. The initial learning rate depends on the step as well as the backbone and is detailed in the supplementary material.

\textbf{Spatial pretraining.} To perform the pretraining of the spatial transformer, we follow SCE \cite{denize2023b}. More specifically we use a projector for the online and target branches and a predictor for the online branch. For data sampling, all sub-videos of 1 second are used. Details can be found in the supplementary material.

\textbf{Spatio-temporal pretraining.} To perform the pretraining of the spatial and temporal transformers, we follow SCE \cite{denize2023b}. We apply a projector on top of each output temporal token and we distill information from the reduced Baidu features. Details can be found in the supplementary material. For data sampling, we randomly extract 150 videos of 32 seconds, or 64 frames, per match at each epoch.

\textbf{Fine-tuning.} To perform fine-tuning, common data augmentations are applied as well as mixup \cite{zhang2018b}. For data sampling, 100 videos per match per epoch are sampled uniformly. The classifier is first initialized and then the whole backbone is fine-tuned. Details can be found in the supplementary material.

\textbf{Inference.} During inference, a sliding window with half overlap is applied on all videos. For multiple timestamp classifications, the maximum of predictions per action is kept. No data augmentation is applied. A hard Non-Maximum Suppression (NMS) of a 5-second window is applied. The 6 first and last seconds of each window are ignored to keep predictions with past or future context.

\subsection{Ablation study}\label{subsec:ablation}

In this subsection, we will make various ablations to highlight the advantages of our 3-step approach, our masking strategy, and how the model and data sampling affect performance. The majority of the ablation study is performed on ViViT Tiny to reduce the cost of training. The models in this section are trained on the training split and evaluated on the validation split.

\begin{table}[]
    \centering
    \small
    \begin{tabular}{lccc}
        \toprule
        Model & Params (M) & GFLOPs & t-AmAP (\%) \\ \midrule 
        ViViT T & 7.5 & 41.2 & 64.7 \\
        ViViT S & 29.1 & 149.5 & 65.9 \\
        ViSwin T & 55.9 & 145.6 & \textbf{66.1} \\ \bottomrule
    \end{tabular}
    \caption{Influence of the model architecture on the t-AmAP.}
    \label{tab:architecture}
\end{table}

\begin{table}[]
    \centering
    \small
    \begin{tabular}{cccc}
        \toprule
        Depth & Params (M) & GFLOPs & t-AmAP (\%) \\ \midrule 
        4 & 7.5 & 41.2 & 64.7 \\
        6 & 8.3 & 41.3 & \textbf{65.4} \\
        8 & 9.2 & 41.3 & 65.3 \\ \bottomrule
    \end{tabular}
    \caption{Influence of temporal depth on ViViT-T for the t-AmAP.}
    \label{tab:temporal-depth}
\end{table}

\textbf{Architectures.} We test two architectures for the spatial part, ViT \cite{dosovitskiy2021} and Swin \cite{liu2021}. Because the output embedding dimension of the spatial transformer is the one used for the temporal transformer, the number of parameters increases quadratically with the spatial dimension. As going deeper with Swin increases the token dimension, its output dimension token is large which leads to a larger number of parameters for ViSwin's temporal encoder in comparison with ViViT's. We compare the performance of ViViT and ViSwin in \cref{tab:architecture}. ViSwin Tiny shows an improvement over ViViT Tiny with $+1.4$ percentage points (p.p). However, this improvement comes with a price of about $7.5$ times more parameters. Going from Tiny to Small for ViViT improved by $+1.2$ p.p for about $4$ times parameter. But, its GFLOPs are slightly higher than ViSwin Small suggesting the Swin spatial transformer is more efficient for larger networks. 

We also test a deeper temporal transformer as the majority of computation comes from the spatial transformer by design \cite{arnab2021}. Indeed for ViT, it sees for a global window 6,272 tokens whereas the temporal transformer only has 32. Therefore, besides increasing the number of parameters, the cost of making a deeper temporal transformer is computationally negligible in comparison with a deeper spatial transformer. The baseline is a depth of 4 blocks of attention and we increase it to 6 and 8. The results are reported in \cref{tab:temporal-depth} and show that increasing the temporal depth to 6 increases the t-AmAP by $0.7$ p.p and going deeper decreases performance. In contrast with ViViT applied to action classification  \cite{arnab2021} we increase performance with a deeper temporal transformer probably because action spotting requires modeling more complex temporal dependencies.

\begin{table}[]
    \centering
    \small
    \begin{tabular}{cc}
        \toprule
        Window duration (s) & t-AmAP (\%) \\ \midrule 
        32 & 64.7 \\
        64 & \textbf{66.0} \\
        128 & 65.5 \\ \bottomrule
    \end{tabular}
    \caption{Influence of temporal length on ViViT-T on the t-AmAP.}
    \label{tab:context}
\end{table}

\textbf{Size of context.} Intuitively, the size of the temporal context influences how our model perceives actions. We study this influence in \cref{tab:context} by increasing $2$ times and $4$ times the temporal context. To keep the computational cost the same between different sizes, we adapt the batch size adequately.  Increasing it $2$ times improved the results by $+1.3$ p.p and a larger context shows a slight decrease. This verifies that for a better understanding of soccer actions, a large temporal context is necessary.

\begin{table}[]
    \centering
    \small
    \begin{tabular}{ccc}
        \toprule
        $\alpha_1$ & $\alpha_2$ & t-AmAP (\%) \\ \midrule
        \multicolumn{3}{l}{\emph{None}} \\
        0.00 & 0.00 & 64.7 \\ \midrule
        \multicolumn{3}{l}{\emph{Only fine-tuning}} \\
        0.00 & 0.50 & 55.5 \\\midrule
        \multicolumn{3}{l}{\emph{Only pretraining}} \\
        0.25 & 0.00 & 64.9 \\
        0.50 & 0.00 & 65.1 \\
        0.75 & 0.00 & 64.8 \\ \midrule
        \multicolumn{3}{l}{\emph{Both}} \\
        0.25 & 0.25 & \textbf{65.2} \\
        0.50 & 0.50 & 65.0 \\
        0.75 & 0.75 & 63.5 \\ \bottomrule
    \end{tabular}
    \caption{Influence of masking ratio during spatio-temporal pretraining ($\alpha_1$) and fine-tuning ($\alpha_2$) on ViViT-T on the t-AmAP.}
    \label{tab:mask}
\end{table}

\textbf{Masking.} Steps 2 and 3 of our training pipeline incorporate a temporal masking strategy. This masking has two goals: limit the overfitting of our model and make the temporal transformer focus on contextual information instead of just aligning its output with its input. We show the advantage of this masking strategy in \cref{tab:mask} by masking only during pretraining, only during fine-tuning, or both. First, masking during only fine-tuning drastically decreases performance by $-9.2$ p.p. Masking during pretraining increases results by up to $0.4$ p.p for 50\% tokens masked and masking during both steps increases up to $0.5$ p.p for 25\% tokens masked. Performance decreases with further masking. These results suggest it is necessary to initialize the mask token during pretraining. Also, the percentage of tokens to mask seems to be different for optimal performance during pretraining and fine-tuning, and fewer masking during fine-tuning seems better.

\begin{table}[]
    \centering
    \small
    \begin{tabular}{cccc}
        \toprule
        Step 1 & Step 2 & Step 3 epochs & t-AmAP (\%) \\ \midrule
        x & x & 100F & 48.1 \\
        SN & x & 100F & 54.7 \\
        IN & x & 100F & 57.7 \\ \midrule
        SN & \checkmark & 50F & 62.2 \\
        x & \checkmark & 30C + 20F & 60.0 \\
        SN & \checkmark & 30C + 20F & 64.7 \\
        IN & \checkmark & 30C + 20F & \textbf{65.0} \\ \bottomrule
    \end{tabular}
    \caption{Influence of the pretraining steps and the number of fine-tuning epochs on ViViT-T on the t-AmAP. \emph{SN} stands for SoccerNet-v2 MoCo self-supervised pretraining, and \emph{IN} for ImageNet21k supervised pretraining. \emph{C} stands for training the classifier and \emph{F} for fine-tuning the whole model.}
    \label{tab:steps}
\end{table}

\textbf{Steps.} Our training pipeline consists of three steps, each adding complexity. Here, we validate the usefulness of each step. We evaluate the quality of our learned representation in step 1 by comparing its performance with a supervised pretrained ViT Tiny model on ImageNet 21k that contains 14 million labeled diverse images. To ensure a fair comparison with spatiotemporal pretrained backbones, when step 2 is not performed, we perform a longer fine-tuning.

We report results in \cref{tab:steps}. Each step consistently improves performance. Indeed, training from scratch reaches $48.1$\% t-AmAP. Adding step 1 increases up to $54.7$\% for SSL pretraining and $57.7$ \% for ImageNet pretraining. This suggests that the spatial transformer takes advantage of initialization from a large diversity of data and that our SSL pretraining can be improved.  Step 2 further improves results, even with random spatial initialization which reached $60.0$\% t-AmAP. With SoccerNet weights, it increases to $64.7$\%, and with ImageNet weights, it reaches $65.0$\%. The gap between SoccerNet and Imagenet pretraining in step 1 is almost closed in step 2. As our SSL approach was trained on videos, while ImageNet weights were obtained on images, we argue that our second pretraining stage benefits from having initial spatiotemporal features. Initializing the temporal transformer accelerates convergence and improves results compared to training from scratch or step 1. This reduces the cost of pretraining, allowing future work to perform fast experiments in the fine-tuning phase.

\begin{table}[]
    \centering
    \small
    \begin{tabular}{cccc}
        \toprule
        Depth & Sequence & Masking & t-AmAP (\%) \\ \midrule
        x & x & x & 64.7 \\
        \checkmark & x & x & 65.4 \\
        x & \checkmark & x & 66.0 \\
        x & x & \checkmark & 65.2 \\
        \checkmark & \checkmark & x & 66.1 \\
        x & \checkmark & \checkmark & 66.2 \\
        \checkmark & \checkmark & \checkmark & \textbf{66.6} \\ \bottomrule
    \end{tabular}
    \caption{Influence of best parameters for temporal depth and length, and the masking strategy on ViViT-T on the t-AmAP.}
    \label{tab:all}
\end{table}

\textbf{All together.} In \cref{tab:all}, we test adding together the different best hyperparameters for a deeper temporal transformer, larger temporal context, and temporal masking. Previously, we showed that the larger improvement came from increasing the temporal context so we add other components to it. Increasing temporal depth adds $0.1$\% whilst using the masking strategy adds $0.2$\% which makes them marginal in comparison with previous improvements. However, combining the three improves $0.6$\% to attain our best result of $66.6$\%. This confirms that a large temporal context is the most determining component to improve performance and that the masking strategy scales with the number of parameters and ensures new information is learned.

\begin{table}[]
    \centering
    \small
    \begin{tabular}{cccc}
        \toprule
        \multirow{2}{*}{Features} & \multicolumn{2}{c}{Pretraining} & \multirow{2}{*}{t-AmAP (\%)} \\
        & Dataset & Fine-tuned & \\ \midrule
        SCE \cite{denize2023b} & K400 &            & 63.6 \\
        SCE \cite{denize2023b} & K400 & \checkmark &  65.7 \\
        Baidu \cite{zhou2021} & K400 & \checkmark & \textbf{66.6} \\ \bottomrule
    \end{tabular}
    \caption{Influence of features to perform KD on ViViT-T on t-AmAP. Supervised features provide best results and self-supervised features of SCE achieve good performance.}
    \label{tab:features}
\end{table}

\textbf{Bank of features.} We change the bank of features used from Baidu \cite{zhou2021}, which necessitates fine-tuning of 5 models pretrained on Kinetics400 \cite{kay2017} to obtain, with two options: extracted features from SCE \cite{denize2023b} pretrained R3D50 on Kinetics400, and its fine-tuned version to the action spotting task. The clips used for the R3D50 last 4 seconds and fine-tuning is performed on the middle frame. We report results in \cref{tab:features}. Baidu features achieve best performance thanks to its 5 aggregated models, but SCE fine-tuned, which is 1 model, is enough to achieve competitive performance. Also, using the self-supervised model loses $-2.1 p.p$ but opens an interesting perspective toward pretraining a self-supervised model on a closer domain for feature extraction.

\begin{table}[]
    \centering
    \small
    \begin{tabular}{cccc}
        \toprule
        NMS & Ignore (s) & Window (s) & t-AmAP (\%) \\ \midrule
        \multicolumn{4}{l}{\emph{Default inference}} \\
        hard & 6 & 5 & 66.6 \\ \midrule
        \multicolumn{4}{l}{\emph{Best inference for soft and hard NMS}} \\
        hard & 12 & 3 & 67.1 \\
        soft & 12 & 10 & \textbf{68.0} \\ \bottomrule
    \end{tabular}
    \caption{Best inference parameters on ViViT-T on the t-AmAP.}
    \label{tab:inference}
\end{table}

\textbf{Inference pipeline.} The inference has also a huge impact on performance. There are 3 parameters that we take into account: whether to use \emph{hard} or \emph{strong} NMS, the number of seconds to ignore at the beginning and end of each window prediction, and the size of the NMS window. In \cref{tab:inference}, we provide the results of the best parameters that we found for hard and soft NMS which are detailed in the supplementary material and we empirically observe a better performance for soft NMS.

\begin{table}[]
    \centering
    \small
    \begin{tabular}{lcc}
        \toprule
        Method & Input & t-AmAP  (\%) \\ \midrule
        NetVLAD++ \cite{giancola2021} & F & 11.5 \\
        AImageLab RMSNet \cite{tomei2020} & F &  28.8 \\
        Baidu \cite{zhou2021} & F & 47.1 \\
        Faster-TAD \cite{chen2022} & F & 54.1 \\
        SpotFormer \cite{cao2022} & F & 60.9 \\
        E2E-Spot \cite{hong2022} & I & 61.8 \\
        Spivak \cite{soares2022} & F & 65.1 \\ \midrule
        COMEDIAN (ViViT-T) & I & 70.7 \\
        COMEDIAN (ViSwin-T) & I & 71.6 \\
        COMEDIAN (ViViT-T - ens.) & I & 72.0 \\ 
        COMEDIAN (ViSwin-T - ens.) & I & \textbf{73.1} \\ \bottomrule
    \end{tabular}
    \caption{Comparison with the state of the art on the test split of SoccerNet-v2. \emph{F} stands for methods using a feature extractor, \emph{I} for methods end-to-end with image inputs.}
    \label{tab:test-sota}
\end{table}

\begin{table}[]
    \centering
    \small
    \begin{tabular}{lcc}
        \toprule
        Method & Input & t-AmAP  (\%) \\ \midrule
        \multicolumn{3}{l}{\emph{Challenge 2022 leaderboard}} \\ 
        Baidu \cite{zhou2021} & F & 49.56 \\
        Transformer-AS \cite{zhu2022} & I & 52.04 \\
        Faster-TAD \cite{chen2022} & F & 64.88 \\
        E2E-Spot \cite{hong2022} & I & 66.73 \\
        Spivak \cite{soares2022} & F & 67.81\\ \midrule%
        \multicolumn{3}{l}{\emph{Challenge 2023 submission}} \\ 
        Spivak* \cite{soares2022} & F & 68.33 \\
        COMEDIAN (ViViT-T - ens.) & I & \textbf{68.38} \\ \bottomrule
    \end{tabular}
    \caption{Comparison with the state of the art on the challenge split of SoccerNet-v2. \emph{F} stands for methods using a feature extractor, \emph{I} for methods end-to-end with image inputs.}
    \label{tab:challenge-sota}
\end{table}

\subsection{Comparison with the State of the Art}

\textbf{Implementation details.} For comparison with the state of the art, we take the best settings found in the ablation study for fine-tuning and inference. The results labeled \emph{ens.} means we use the average predictions of 3 seeds. We evaluate on the test split as well as the challenge split. When we evaluate on the test split, spatiotemporal pertaining and fine-tuning are performed on the training and validation splits, and for the challenge all annotated splits are used.

\textbf{Comparison on test split.} We report our results in \cref{tab:test-sota}. We compare ourselves with methods that use a sequence of images as input or a feature extractor. We observe that COMEDIAN with ViViT Tiny provides a significant improvement over the state of the art by $+5.6$ p.p on t-AmAP. ViSwin Tiny increases performance by $+0.9$ p.p but at a high cost in terms of computational usage. Finally using an ensemble of our 3 seed, we achieve $72.0$\% t-AmAP for ViViT Tiny and $73.1$ \% for ViSwin Tiny. These results empirically prove that our approach even with a small network produces state-of-the-art results by using our initializing pipeline. It is worth noting that we perform a simple fine-tuning stage. Previous approaches only focused on the fine-tuning part and because the two are not mutually exclusive, it opens interesting perspectives for future work to build better fine-tuning upon our approach.

\textbf{Comparison on challenge split.} We report our results in \cref{tab:challenge-sota}. We compare with some participants from the Challenge 2022. We also report the result of the baseline of the Challenge 2023 \cite{soares2022} in which we participated, which is an improved version of the winner of 2022. Our proposed COMEDIAN achieves $68.38$\% on global t-AmAP for ViViT Tiny. Contrary to the test split we do not have a gap with previous state-of-the-art methods and achieve $+0.56$ p.p in comparison with 2022 best result and $+0.05$ p.p in comparison with 2023's. Because of the opacity of the challenge split's labels, it is difficult to investigate the discrepancy between the test and the challenge. Compared with the best end-to-end methods E2E-Spot \cite{hong2022}, our approach achieves a more significant improvement of $+1.64$ p.p on t-AmAP.

\section{Discussion}\label{Discussion}
We discuss potential improvements and future directions for enhancing COMEDIAN. For spatial pretraining, we identify three directions: increasing the amount of training data, the number of frames, and designing a SSL task for semantically sparse images. For spatio-temporal pretraining, we suggest exploring temporal models better suited to action spotting and improving KD by using SSL-learned features from a large close domain.  In the fine-tuning stage, we suggest improving the labelization and designing losses that make use of relationships between different action classes and are capable of dealing with challenging actions such as card-related ones. Finally, our study focused on soccer action spotting and studying the generalization to other datasets and general TAD is an interesting perspective.

\section{Conclusion}\label{Conclusion}

In this chapter, we introduce COMEDIAN a novel approach for Action Spotting that leverages self-supervised learning and knowledge distillation to initialize a spatio-temporal transformer. It achieves state-of-the-art results on the SoccerNet-v2 action spotting task, demonstrating the effectiveness of the proposed pipeline. By utilizing unlabeled video data for pretraining, we address the subjective and resource-intensive manual labeling processes for action spotting. The pretraining cost is leveraged by a faster and better convergence during fine-tuning. While our approach shows promising results, there are areas for improvement in the pretraining and fine-tuning steps and we hope that our approach will open the path to new methods to increase performance on action spotting and temporal action detection with spatio-temporal transformer models.


\section*{Acknowledgement}\label{Acknowledgement}
This publication was made possible by the use of the Factory-AI supercomputer, financially supported by the Ile-de-France Regional Council, and the HPC resources of IDRIS under the allocation 2023-AD011014382 made by GENCI.

{\small
\bibliographystyle{ieee_fullname}
\bibliography{arxiv}
}

\clearpage

\appendix

\section{Implementation details}

\begin{table*}[th]
    \centering
    \small
    \begin{tabular}{l|c|c|c} \toprule
     & ViViT Tiny & ViViT Small & ViSwin Tiny\\ \midrule
    Input dim &  $C \times T_g \times H \times W$ & $C \times T_g \times H \times W$ &  $C \times T_g \times H \times W$ \\ \midrule
    \multicolumn{4}{l}{\emph{Spatial encoder}} \\
    Input tokens dim & $\left(\frac{H}{16} \times \frac{W}{16} + 1\right) \times \frac{T_g}{2} \times 192$ & $\left(\frac{H}{16} \times \frac{W}{16} + 1\right) \times \frac{T_g}{2} \times 384$ & $\frac{H}{4} \times \frac{W}{4} \times \frac{T_g}{2} \times 96$ \\
    Num parameters & $5.7$M & $22.0$M & $27.5$M \\
    GFLOPs & $41.19$ & $149.30$ & $144.68$ \\ \midrule
    \multicolumn{4}{l}{\emph{Temporal encoder}} \\
    Input tokens dim & $\frac{T_g}{2} \times 192$ & $\frac{T_g}{2} \times 384$ & $\frac{T_g}{2} \times 768$ \\
    Num parameters & $1.8$M & $7.1$M & $28.4$M \\
    GFLOPs & $0.06$ & $0.24$ & $0.96$ \\ \midrule
    \multicolumn{4}{l}{\emph{Global model}} \\
    Num parameters & $7.5$M & $29.1$M & $55.9$M \\
    GFLOPs & $41.25$ & $149.54$ & $145.64$ \\
     \bottomrule
    \end{tabular}
    \caption{Comparison of the ViViT Tiny, ViViT small, and ViSwin Tiny spatial and temporal encoders and global model in terms of computational usage.}
    \label{tab:stats}
\end{table*}

\subsection{Architectures}

\textbf{Encoders.} For the spatial encoder, two different transformer architectures are used: ViT \cite{dosovitskiy2021} and Swin \cite{liu2021}. By default, the temporal encoder is a stack of 4 attention layers as in ViT architecture. For ViT, the global architecture corresponds to the ViViT model 2 \cite{arnab2021}. We keep this name and refer to the Swin based-architecture as ViSwin. In \cref{tab:stats}, we provide the input dimension of tokens for the spatial and temporal encoders, and their number of parameters and GFLOPs for ViViT Tiny, ViViT Small, and ViSwin Tiny.

For all models, the majority of the computations are performed in the spatial encoder which sees a lot of tokens, and the temporal encoder computational cost is negligible. However, the number of parameters does not scale well with the output dimension of the spatial encoder, due to the self-attention mechanism, which is reflected in ViSwin Tiny. It has $4$ times more temporal parameters than ViViT small but only $1.25$ times more spatial parameters. However, as Swin has less reduced computational usage in comparison with ViT by design \cite{liu2021} it scales better to deeper spatial architectures.

\subsection{Optimizers}

We use the optimizer ADAMW for pretraining and finetuning with a weight decay of $0.05$. The initial learning rate depends on the training step as well as the backbone as detailed below. However, the steps follow different linear scaling rules for an initial learning rate $\eta$:
\begin{itemize}
    \item Step 1: $\eta_{scaled}= \eta \times \frac{batch\_size}{256}$
    \item Step 2 and 3: $\eta_{scaled}= \eta \times \frac{batch\_size}{256} \times \frac{T_g}{64}$ with $ T_g$ the number of global frames per video.
\end{itemize}

\textbf{Step 1.} The initial learning rate is $\eta = 5\times10^{-4}$ with 10 epochs of warmup and a cosine annealing scheduler is applied throughout training.

\textbf{Step 2.} The initial learning rate is $\eta = 0.002$ with 10 epochs warmup and cosine annealing scheduler that ends at $0.01 \times \eta$.

\textbf{Step 3.} The initial learning rate is $\eta = 5\times10^{-4}$ for ViViT and $\eta = 3\times10^{-4}$ for ViSwin that ends at $0.01 \times \eta$.

\subsection{Spatial Pretraining.}

To perform the pretraining of the spatial encoder, we follow practices introduced by $\rho$MoCo \cite{Feichtenhofer2021} and SCE \cite{denize2023b}. More specifically we use a 3 layers Multi-Layer Perceptron (MLP) on top of the online and target encoders of hidden size $1024$ and output size $256$ that is discarded after this step. The online predictor is a 2 layers MLP with the same hidden and output size as the projectors. The data augmentation distributions are the standard contrastive ones used on images \cite{denize2023b} and the temperature applied is $\tau = 0.1$. The momentum buffer size is $65,536$. For data sampling, all sub-videos of 1 second, or 2 frames, are used. The model is trained for 100 epochs with a batch size of $1024$. 

\subsection{Spatio-temporal pretraining.} 

To perform the pretraining of the spatio-temporal encoder, we follow practices introduced by SCE \cite{denize2023b}. More specifically we use a 3-layer MLP on top of the temporal encoder of hidden size $1024$ and output size $512$ to match the dimension of the Baidu \cite{zhou2021} features. The projector is later discarded. For the SCE loss parameters we use $\tau = 0.1$, $\tau_m = 0.07$, $\lambda = 0.5$. The data augmentation used is the $strong_\gamma$ without cropping reported in the SCE paper. For data sampling, we randomly extract 150 videos of 32 seconds or 64 frames per game at each epoch. The batch size for 32-second videos at 2 FPS is 64. For longer clips, the batch size is inversely proportional to the length and number of windows. For example, for 64 seconds, the batch size is 32 and the number of windows sampled per match is 75. 

\subsection{Finetuning.} 

A linear classifier is applied to each output temporal token to perform fine-tuning. Each video sampled is augmented by using color jittering with probability $0.8$ and of strength $\pm 0.4$ on brightness, contrast, and saturation and $0$ for hue to avoid changing the color of cards. Random Gaussian blur is also applied with probability $0.5$ and a kernel size of $23$ with $\sigma \in [0.1, 2.]$. A horizontal flip of probability $0.5$ is also applied followed by a mixup \cite{zhang2018b} whose mixing coefficient is sampled by a Beta law $\mathcal{B}(0.1, 0.1)$. 

The classifier is first trained during 30 epochs for its initialization and then the whole architecture is fine-tuned for 20 epochs for ViViT and 10 for ViSwin. The learning rate is reset for the second part.

For data sampling, 100 videos per match are uniformly sampled whilst enforcing that the beginning and end of each half are selected to avoid missing kickoffs and last-second actions. The batch size for 32-second videos at 2 FPS is 128. For longer clips, the batch size is inversely proportional to the length and number of windows. For example, for 64 seconds, the batch size is 64 and the number of windows sampled is 50 per match.

\section{Inference hyper-parameters search}

\begin{table}[t]
    \centering
    \small
    \begin{tabular}{cc} \toprule
         Seconds & t-AmAP (\%) \\ \midrule
         0 & $66.4$  \\
         2 & $66.6$ \\
         4 & $66.7$ \\
         6 & $66.6$ \\
         8 & $66.6$ \\
         10 & $\mathbf{66.8}$ \\
         12 & $\mathbf{66.8}$ \\
         14 & $\mathbf{66.8}$ \\ 
         16 & $66.7$ \\ \bottomrule
    \end{tabular}
    \caption{Influence of the number of seconds ignored at the start and end of each window prediction on the t-AmAP.}
    \label{tab:inf-ignore}
\end{table}

\begin{figure}[]
    \includegraphics[width=1.\linewidth]{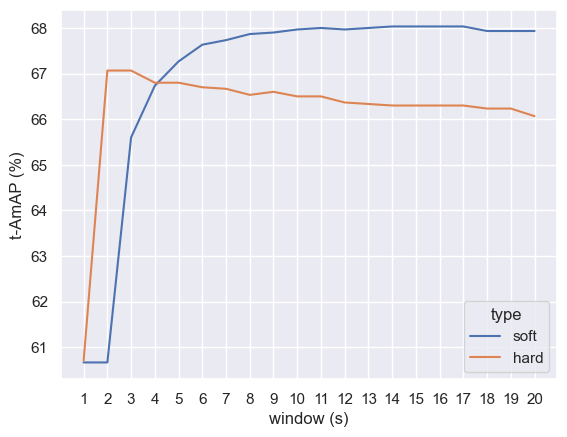}
    \caption{Influence of the soft and hard NMS and its window size in second on the t-AmAP.}
    \label{fig:inf-nms}
\end{figure}

During inference, a sliding window with half overlap is applied on all videos. For multiple timestamp classifications, the maximum of predictions per action is kept. No data augmentation is applied. By default, a hard Non-Maximum Suppression (NMS) of a 5-second window is applied. The 6 first and last seconds of each window prediction are ignored to keep predictions that have past and future context.

In \cref{tab:inf-ignore}, we study the effect of varying the number of seconds to ignore. Taking all predictions has the worst result of $66.4$\% t-AmAP showing that it is interesting to remove predictions on edge that do not have access to the context from the past or the future. The results increase up to $66.8$\% at 10 seconds and are stable for further seconds ignored. The increase in performance is relatively low and can be explained by the fact that the inference sliding window allows for some undetected predictions on edges to be retrieved by past or future windows.

In \cref{fig:inf-nms}, we study the effect of using Hard or Soft NMS. As for \cite{soares2022}, we see an increase in using soft NMS over hard NMS. Depending on the NMS type the optimal temporal window size for NMS is not the same. The best results are achieved for a hard NMS with a 4-5 seconds window at $66.8$\% t-AmAP and $68.0$\% for a soft NMS with an 11-17 seconds window. The results show that not only does soft NMS perform better than hard NMS but is also more stable.

\clearpage

\end{document}